\documentclass[letterpaper, 10 pt, conference]{ieeeconf}  

\usepackage{amsmath}
\usepackage{amssymb}
\usepackage{graphicx}
\usepackage{tikz}
\usepackage{comment}
\usetikzlibrary{shapes.geometric}

\usepackage{subcaption}

\IEEEoverridecommandlockouts                              

\title{\LARGE \bf
Neural-Guided Runtime Prediction of Planners for Improved Motion and Task Planning with Graph Neural Networks
}

\author{Simon Odense$^{1}$, Kamal Gupta$^2$, and William G. Macready$^{3}$
\thanks{$^{1}$ Sanctuary Cognitive Systems Corp. and School of Engineering Science, Simon Fraser University. {\tt\small simon\_odense@sfu.ca}}%
\thanks{$^{2}$ School of Engineering Science, Simon Fraser University, Burnaby, BC Canada. {\tt\small kamal@sfu.ca}}%
\thanks{$^{3}$ Work performed at Sanctuary Cognitive Systems Corp., Vancouver, BC, Canada. Current email: {\tt\small william.macready@relational.ai}}%
}

\begin{document}

\maketitle
\thispagestyle{empty}
\pagestyle{empty}

\begin{abstract}
The past decade has amply demonstrated the remarkable functionality that can be realized by learning complex input/output relationships. Algorithmically, one of the most important and opaque relationships is that between a problem's structure and an effective solution method. Here, we quantitatively connect the structure of a planning problem to the performance of a given sampling-based motion planning (SBMP) algorithm. We demonstrate that the geometric relationships of motion planning problems can be well captured by graph neural networks (GNNs) to predict SBMP runtime. By using an algorithm portfolio we show that GNN predictions of runtime on particular problems can be leveraged to accelerate
online motion planning in both navigation and manipulation tasks. Moreover, the problem-to-runtime map can be inverted to identify subproblems easier to solve by particular SBMPs. We provide a motivating example of how this knowledge may be used to improve integrated task and motion planning on simulated examples. These successes rely on the relational structure of GNNs to capture scalable generalization from low-dimensional navigation tasks to high degree-of-freedom manipulation tasks in 3d environments. 

\end{abstract}

\section{INTRODUCTION}

Motion planning, a fundamental problem in robotics, seeks a collision-free path from an initial robot state to a given goal state. One of the most successful classes of algorithms for this problem is Sampling Based Motion Planners (SBMP) in which random configurations of the robot are generated and paths (or edges) between two sample configurations are generated using a collision checker to ensure that they are collision-free. The resulting graph is a discrete representation of the free configuration space (c-space) of the robot, and a successful path search through this graph yields a collision-free path from the initial to the goal configuration of the robot \cite{essentials}. There are many SBMP variants such as RRT \cite{RRT}, RRTConnect \cite{RRTConnect}, and PRM \cite{journals/trob/KavrakiSLO96} which have different strengths and weaknesses. They all, however, share the same fundamental characteristic of relying on randomly generated states to construct a path. Despite theoretical guarantees of eventual convergence for many of the SBMPs, their random nature makes it difficult to determine how successful a motion planner will be for a given problem.

Recently, neural networks have been used to address some of these deficiencies in what may be called Neural Guided Motion Planning (NGMP) -- see Section \ref{sec:related} for a brief survey of NGMP. Most of these techniques have focused on using neural networks to learn good configuration space sampling distributions for SBMPs. Much less work has been done on meta-level NGMP in which neural networks are used to predict the effectiveness of a given SBMP for a given problem. We address this by using neural networks for the speed prediction problem where we predict expected completion time of a SBMP for a given motion planning problem. We do this by training graph neural networks (GNNs) on graphical representations of both 2D navigation problems and high degree-of-freedom motion planning problems for manipulator arms in 3D scenes within the iGibson simulation environment for robot training \cite{shen2021igibson}. 
Conversely, we also show that GNNs can accurately predict the problem with the lowest expected completion time among randomly generated sets of test problems for a given SBMP.

An important potential application for this latter prediction is Task And Motion Planning (TAMP) where  a robot is required to solve a high-level task such as stacking objects. A solution to a TAMP problem consists of a sequence of discrete \textit{actions} which complete the objective. Each action must be completed with a feasible continuous motion. Coordinating the high-level discrete task planning task with the low-level motion planning is a challenging problem. Often, a hierarchical approach is taken in which a task planner generates a sequence of actions and IK-solvers and motion planners are used to generate trajectories which implement these actions \cite{DBLP:journals/corr/abs-2010-01083}. Accurate prediction of runtimes would be helpful for efficiently solving TAMP problems. The lack of \textit{a priori} knowledge regarding the expected completion time of a motion planner can result in significant time loss as the failure of a motion planner to solve a subproblem within a given amount of time will trigger re-planning at the task level until sufficiently simple motion planning problems are generated by the task planner. Conversely, the knowledge that a problem is easily solved by a given SBMP can be used to guide an appropriate sequence of actions at the the task planning level.

In this paper we explore several related questions:
\begin{enumerate}
    \item can we predict the runtimes of SBMP solvers from the characteristics of the planning problem?
    \item can the geometric relationships inherent in planning problems be leveraged to make such predictions scalable?
    \item can a quantitative understanding of the characteristics/runtime relationship be exploited to improve upon existing motion planners by using portfolios of solvers?
    \item can runtime prediction also improve task and motion planning?
\end{enumerate}
Our key contribution is to demonstrate affirmative answers to all these questions.

We begin in subsection \ref{sec:related} by situating our contributions within the context of recent work combining neural models and motion planners. With this background, in section \ref{sec:MODEL} we develop a graph neural network (GNN) model reflecting the geometric structure of planning problems and use this as a foundation to address the above questions in section \ref{sec:RESULTS}. Firstly, in section \ref{sec:portfolios} we validate the performance of GNNs as classifiers on a navigation task to enable an accurate problem-dependent prediction permitting the construction of SBMP algorithm portfolios. Section \ref{sec:probPredicton} then extends the GNN classification to predict algorithm runtimes and shows that we can \textit{a priori} identify the hardness of a problem for a given planner. In this section we then sketch how this ability might be used to aid in a stylized example of task and motion planning. Lastly, in section \ref{sec:manipulation} we demonstrate that GNNs, unlike occupancy grid methods, scale well to higher-dimensional planning problems by considering an arm manipulation task in a realistic setting.

\section{Related Work}
\label{sec:related}

Given the pressing need for fast motion planning, recent research has explored the use of 
neural models trained offline to accelerate online planning. In the context of SBMPs, a natural approach to this, and the one that has received the most attention, is to use neural networks to learn a configuration sampling distribution which builds a well-tuned search graph that focuses search effort on regions likely to be on the solution path. To do so, examples of successful motion planning paths are generated offline to train a neural network to model a conditional distribution over the configuration space. Most commonly, these models condition on a problem description in the voxelized form of an occupancy grid or point-cloud \cite{CVAE1}\cite{CVAE2}\cite{NGMP2}.\footnote{ We learn from these works to avoid discretization of the geometry (as in occupancy grids) and adopt a sparse relational approach which scales better with dimensionality.
} Pleasingly, with this approach, configuration samples may be passed into \emph{any} SBMP to construct the graph in the usual fashion. This permits the application of this method to a wide variety of low-dimensional problems that may be more amenable to one SBMP or another. Our work is orthogonal to this neural sampling approach, and may be combined with it.

Other approaches rely on the neural network to directly predict a path. This can be done by training a neural network to predict the next point in a solution path given the current one \cite{DBLP:journals/trob/QureshiMSY21}. These methods forgo SBMP algorithms and instead use the predictions of the neural network directly along with some additional logic (although they may utilize SMBP algorithms as backup planners). Again, such methods may be combined with our work.

Another avenue for neural-guided motion planning that has received less attention is what may be referred to as meta-level neural-guided motion planning. In meta-level neural-guided motion planning, neural networks are used to make predictions \textit{about} a motion planning algorithm. This generally involves predicting the performance of an algorithm for a given problem. \cite{DBLP:conf/iros/SungKL21} has explored the optimal stopping time problem for anytime motion planners. In this problem, neural networks predict the time at which an anytime motion planner (in this case RRT* \cite{RRT*}) should stop running in order to achieve the optimal trade off between solution quality and time according to a weighted score function. In this case, it is assumed that the algorithm has already found a solution and the question being addressed is determination of the continued runtime to improve the solution. In our work, we address a different meta-level problem, namely,  how long it will take an SBMP to find a solution for a given problem and how might we exploit this information? Such, performance prediction has been addressed in other domains \cite{10.5555/2832747.2832840} and we explore its use in SBMP through problem-specific selection of algorithms from within a portfolio of SBMP algorithms.

\section{THE MODEL}\begin{figure}[t]
\centering
\includegraphics[scale=0.8]{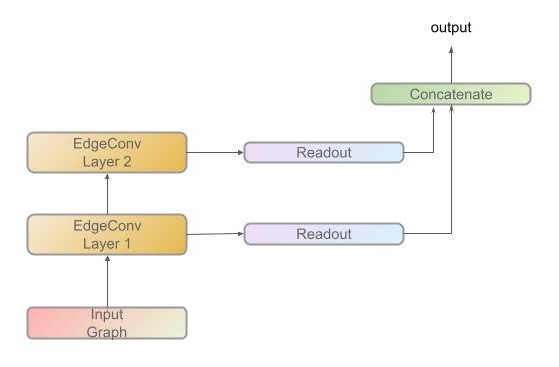}
\caption{A 2-Layer EdgeConv GNN. Each EdgeConv layer applies Eq.~\eqref{eq:edgeconv} to its input graph to produce an output graph. The Readout function is applied to the output of each layer according to Eq. ~\eqref{eq:readout} and the result is concatenated to obtain the output  } 
\label{fig:EC_arch}
\end{figure}
\label{sec:MODEL}

\begin{figure*}[t]
  \centering
  \begin{subfigure}{0.4\textwidth}
  \includegraphics[scale=0.5]{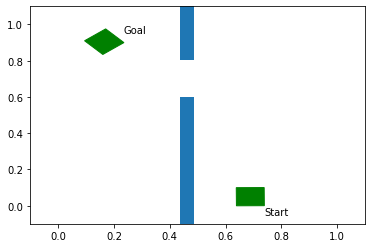}
  \caption{An example problem from the 2D dataset}
   \end{subfigure}
 \begin{subfigure}{0.4\textwidth}
 \includegraphics[scale=0.23]{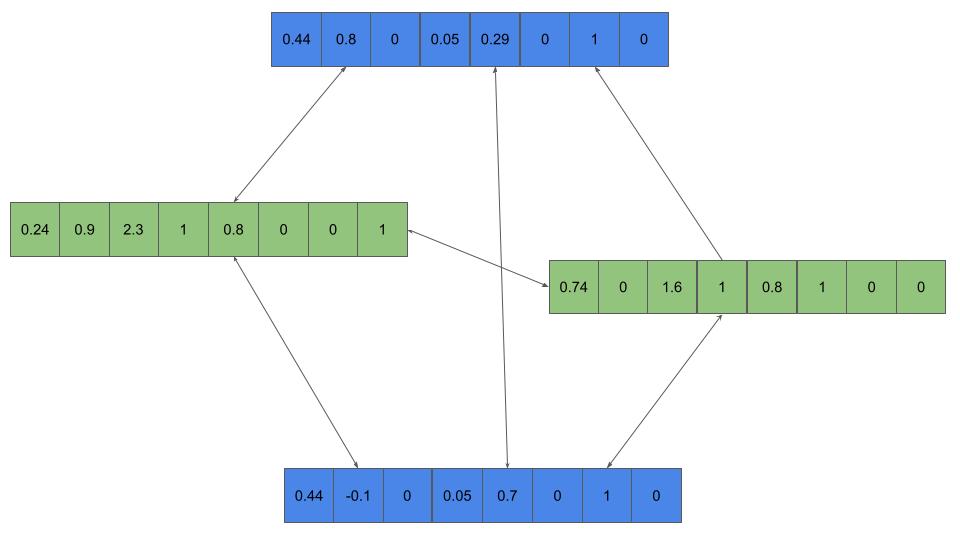}
\caption{Graph representation of problem (a)}
\end{subfigure}
\caption{A sample navigation problem from the 2D dataset with the corresponding graph representation used as input for the GNN. Node features include the position, orientation (in radians), width, height, and a one-hot label indicating the object type; i.e. start, obstacle, or goal.}
\label{fig:2D_examp}
\end{figure*}
The structure of a motion planning problem is relational. Objects in the environment are well-defined and have clear geometric relationships with each other and with the robot. The robot itself is naturally described with a transform tree of related coordinate frames. These relationships may be captured in a graph where nodes represent objects or parts of the robot and where labelled edges indicate pairwise geometric relationships. To reflect this structure for effective generalization, we rely on a graph neural network (GNN) architecture. GNNs are a type of neural network that take graph input and generalize convolutional neural networks with filters that act on each graph node using information from neighbouring nodes \cite{GNN1}\cite{GNN2}. 


Each node $v$ (and possibly the edges) in the input has an associated input feature vector, $h_v$, which is refined through a layerwise mapping with neighbouring feature vectors. The outputs of the $k^{th}$ layer of a GNN are given by the following general formulation \cite{DBLP:conf/iclr/XuHLJ19}:
\begin{equation}
\begin{split}
    a^{(k)}_v & =AGGREGATE^{(k)}(\{h^{(k-1)}_u : u \in N(v)\}) \\
    h^{(k)}_v & =  COMBINE^{(k)}(h^{(k-1)},a^{(k-1)})
    \end{split}
\end{equation}
where $N(v)$ is the set of neighbours of $v$. GNNs may also be used to learn global properties, $h_G$, of the entire graph. In this case, node features for a layer are aggregated using a $READOUT$ function as follows
\begin{equation}
    h_G^{(k)}= READOUT(\{h^{(k)}_v : v \in G \}).
    \label{eq:readout}
\end{equation}
GNNs have had application to diverse fields such as chemistry, computer vision, and knowledge graph completion and many different architectures have been developed \cite{zhou2018graph}. For our experiments, we use the EdgeConv architecture \cite{DBLP:journals/tog/WangSLSBS19}.

In EdgeConv, the \textit{AGGREGATE} and \textit{COMBINE} operations are combined into the following:
\begin{equation}
h^{(k)}_v=\sum\limits_{u\in N(v)} \theta^{(k)} (h^{(k-1)}_v,h^{(k-1)}_u-h^{(k-1)}_v).
\label{eq:edgeconv}
\end{equation}
Here, $\theta^{(k)}$ is a Multi-Layer Perceptron (MLP), a type of neural network with multiple layers in which every neuron in one layer is connected to every neuron in the following layer. The MLP takes as input the features of the target node and the difference of features between the target node and its neighbouring node. The EdgeConv architecture is advantageous in that it uses both local and global data to update the features, ideal for geometric problems such as motion planning. For our readout function we use a sum. When considering the output of the whole graph, we apply the readout to each layer in the graph network and concatenate the result. This allows us to capture important features found at each layer of the graph network. Fig.~\ref{fig:EC_arch} provides a visual overview.

To our knowledge, GNNs have seen limited application to motion planning problems. However, several methods using GNNs have been developed and tested on low dimensional problems \cite{GNNmp}. These methods use GNNs by forming graphs from a collection of points in the configuration space of a motion planning problem. Each point and edge is given features such as it's position and whether or not it's occupied. Using these graphs, GNNs were used to predict a variety of targets, such as the set of graph nodes critical for a search algorithm or heuristics that can be used to guide the search.

Unlike the previously discussed examples, we train our GNNs on graph representations of the \textit{workspace} instead of the c-space. The advantage of this is that the geometry of c-space is highly complex and often only implicitly defined whereas sparsity in the workspace is explicit. This allows us to represent 3D motion planning problems quickly with relatively small graphs even when there are high number of degrees of freedom involved. Using sensor readings, a robot can build a graph representing the environment to be used with our GNN model quickly, rather than having to build a potentially complicated model of the configuration space. Updating this graph as the robot moves is simple as the features of existing nodes can be updated to represent the new location of various objects and nodes can be added as the robot approaches new objects.

To apply GNNs to motion planning problems, we represent objects in the workspace as nodes in the graph with connections between nodes added based on spatial proximity. Objects in the workspace are either obstacles to be avoided or links of the robot. Nodes are given features representing the geometric information of the object as well as a label indicating the type of object. Fig.~\ref{fig:2D_examp} gives an example of the graph construction for the 2D navigation dataset (details of features are given in \ref{sec:portfolios}). The graph topology for the EdgeConv network is generated using a k-nearest neighbours approach in which each node is connected to the k neighbours it is spatially closest to. k  is a hyperparameter that is chosen experimentally. For example, in our 2D navigation dataset k was selected to be 3 meaning that in Fig. ~\ref{fig:2D_examp} the graph is fully connected. Future work may experiment with other graph topologies, for instance, in some cases, particular background knowledge on specific problems may be used to hand-design the graph topology.

In order to make a problem prediction, the GNN is applied to the input graph and its output is then fed through a final fully-connected network to produce the final prediction. For 2D navigation problems, the graph representation is sufficient to capture the entire problem information. This is done by representing both the initial and final position of the moving object as nodes in the graph. This method of representation could also be used for our 3D motion planning dataset by including 2 nodes for each robot link, one representing the initial position of the link and one representing the final position of the link. However, to reduce the graph size, as well as to provide the network with additional information about the motion planning problem that could be useful, we instead add nodes for only the initial position of each link and include the initial and goal states of the configuration space in the problem description. When this is the case, we use an additional fully-connected network to process this information. We then use the output of the GNN and the fully-connected network as input into the final fully-connected layers of the network, see Fig.~\ref{fig:nn_arch}. 
Full details of node features and network architecture for each experiment are given in the results section.



\begin{table}[t]
    \caption{Total expected completion time (in seconds) of 1000 test set navigation problems using an SBMP chosen by various predictors.}
\label{table_algoport_2D}
\begin{center}
\begin{tabular}{|c||c|}
\hline
Predictor & Expected Total Runtime\\
\hline
\hline
Perfect  & 368 \\
\hline
\textbf{GNN} & 389 \\
\hline
Fully-connected & 397  \\
\hline 
RRTConnect & 429  \\
\hline
\end{tabular}
\end{center}
\end{table}
\section{RESULTS}
\label{sec:RESULTS}

We present results on two canonical types of motion planning: i) 2D navigation tasks in cluttered environments and ii)  motion-planning for a 7 degree-of-freedom manipulation task in 3D environments. For each type of task we evaluate the computational savings accruing from problem-specific algorithm selection from a portfolio of four SBMPs. Next, we demonstrate that runtime prediction may also be used to identify problems that are likely to be hard for a given algorithm and use this as a proof-of-concept to accelerate task planning. Finally, we turn to higher-dimensional planning for manipulators and demonstrate that the relational structure of GNNs permits scaling to larger problems. The SBMPs we investigate are all RRT-derivatives, but given that our method is agnostic to the choice of algorithm, it could be applied to any SBMP or even motion planning algorithms that are not sampling based. All networks were trained using PyTorch with the PyTorch geometric package used for the GNNs \cite{PyTorch}\cite{PyTorchgeo}.

\subsection{Navigation: portfolios of SBMPs}
\label{sec:portfolios}
The first problem we examine is the prediction of the fastest SBMP for a given problem. Given a set of SBMPs, for each motion planning problem, we want to be able to choose the SBMP with the fastest expected runtime. We test the ability of GNNs to solve this problems against fully-connected networks, a naive, and a perfect predictor on a 2D narrow gap navigation dataset.

These motion planning tasks require moving a block from a starting position to a goal position through a number of narrow gaps. Random problems are generated as follows: The block positions consist of an $x,y$ location in $[0,0.9]\times[0,0.9]$ and a rotation angle, $\theta$ in $(-\pi,\pi]$. The angle and $x$ position of the initial and final states are randomized in each problem while the initial $y$ position is set to $0$ and the final $y$ position is set to $0.9$. One to three vertical barriers stretching the entire height of the workspace are placed at random $x$ positions. Most placements are controlled to allow sufficient room between blocks for feasible navigation although some are left uncontrolled in order to generate a small number of infeasible problems to aid in training. Gaps of various widths are placed in each of these barriers at randomized heights. See Fig.~\ref{fig:2D_examp} for an example problem and feature description. $5000$ random problems are generated and solved with four RRT variants (RRT \cite{RRT}, RRTConnect \cite{RRTConnect}, TRRT \cite{trrt} and lazyRRT \cite{Bohlin01arandomized}) using OMPL \cite{OMPL}. We run each algorithm 40 times (with default parameters) on a given problem and record the average completion time of each algorithm. The timeout is set to 3 seconds.

\begin{table}[t]
\caption{Cumulative expected completion time (in seconds) of SBMPs on 2D navigation problems predicted to be the fastest by various predictors among a randomly chosen set of test problems.}
\label{table_random_problem_2D}
\begin{center}
\begin{tabular}{|c||c||c||c||c|}
\hline
Predictor & RRTConnect & RRT & TRRT & lazyRRT\\
\hline
\hline
Perfect  & 32 & 64 & 102 & 156\\
\hline
\textbf{GNN} & 56 & 90 & 143 & 181 \\
\hline
Fully-connected & 87 & 147 & 189 & 244 \\
\hline
Least obstacles & 381 & 433 & 483 & 1350 \\
\hline
Random & 497 & 536 & 729 & 1406 \\
\hline
\end{tabular}
\end{center}
\end{table}
Different SBMPs are suited to different kinds of motion planning problems, and the variance (across algorithms) in runtime on a given problem can be significant. Thus, in real-time and computation-bound settings there is the potential of speeding planning with a computationally cheap selection of a well-suited SBMP. We could address this problem by regressing algorithm runtime on problem features, but this unnecessarily complicates prediction; we only really require an ordering amongst algorithms, or more simply the best performer from a pre-specified algorithm portfolio. 

We apply GNNs to this problem by training a classifier to predict the optimal algorithm for a given problem. The GNN maps from the learned node features across the entire planning graph to a four-component one-hot label indicating which of (RRT, RRTConnect, lazyRRT, TRRT) has the fastest expected runtime for the problem.
\begin{table}[t]
\caption{Expected cumulative runtime of RRTConnect (in seconds) for solutions to TAMP problems selected by various predictors.}
\label{table_TAMP_problem_2D}
\begin{center}
\begin{tabular}{|c||c|}
\hline
Predictor & RRTConnect \\
\hline
\hline
Perfect  & 1.97 \\
\hline
\textbf{GNN} & 6.13 \\
\hline
Fully-connected & 12.88  \\
\hline
Least obstacles & 61.32  \\
\hline
\end{tabular}
\end{center}
\end{table}
As shown in Fig.~\ref{fig:2D_examp}, the input node features include the $x$ and $y$ position of the object, its rotation angle, width, height and a three-component one-hot label indicating whether the node represents an initial position, an obstacle, or the goal position. We place edges using the 3-nearest neighbours of each node. Our neural network consists of a 2-layer edge convolutional GNN followed by a 2-layer fully-connected network and a final softmax layer. The first edge-convolutional layer uses a 2-layer MLP with $512$ and $512$ neurons respectively, the second edge-convolutional layer uses a 2-layer MLP with $256$ and $256$ neurons respectively. The SELU activation function was used for the hidden nodes. In order to compare the effectiveness of the GNN on this problem to other networks, we trained another neural network predictor using a fully-connected architecture. The input to this network consisted of an occupancy-grid representation of the workspace along with the initial and goal state. 
To generate output, the occupancy grid is fed into a $2$-layer fully-connected network of $512$ and $256$ hidden neurons, the initial and goal state is fed into a $2$-layer fully-connected network of $100$ and $50$ neurons, and finally, the output of each of these is concatentated and fed into a final $2$-layer network of $400$ and $200$ neurons before the output softmax layer. This network architecture was based on the networks used to train a neural sampler in \cite{CVAE1}. A grid search was performed on both the GNN and the fully-connected network to find good hyperparameters. The networks were trained using the ADAM optimizer with negative log-likelihood loss and a 80/20 training/test split on the dataset. We tested the networks by using them to predict the fastest algorithm for each problem in the test set.

Table~\ref{table_algoport_2D} summarizes the expected times of the learned predictors and compares it to other candidates. The perfect predictor is the lowest possible cumulative runtime that selects the best algorithm for each problem. We see that the GNN predictor comes very close to this best-possible cumulative runtime. For these problems, the single fastest planner is RRTConnect which achieved the fastest time in about $80\%$ of examples. Consequently, we show the runtime resulting when RRTConnect is always selected. Even in this unbalanced dataset, the GNN predictor is able to markedly improve upon the dominant RRTConnect. Across a more balanced portfolio (with all solvers similarly comparable) we expect larger benefits from GNN-based SBMP portfolios.

\subsection{Navigation: problem prediction}
\label{sec:probPredicton}
The next experiment we conduct is to examine the ability of a GNN to predict the expected runtime for a single algorithm on a given problem. We measure the effectiveness of this by presenting the trained GNN with a set of test problems and using it to predict the fastest. 
The previous experiments assessed the ability of GNNs to predict the fastest algorithm for a given planning problem. Next, we explore the utility of the inverse problem: given a SBMP method, can we identify problem characteristics that might make it easier to solve? We answer this question affirmatively and suggest an application to task and motion planning (TAMP). We provide a proof-of-concept demonstration that TAMP may be accelerated through offline learning of the problem/runtime relationship. 

\begin{table}[t]
\caption{Cumulative expected completion time (in seconds) of RRTConnect on iGibson (7DOF) motion planning problems predicted to be the fastest by various predictors among a randomly chosen set of test problems.}
\label{table_random_problem_3D}
\begin{center}
\begin{tabular}{|c||c|}
\hline
 Predictor & RRTConnect \\
\hline
\hline
Perfect  & 890\\
\hline
\textbf{GNN} & 1382 \\
\hline
Least obstacles & 1941 \\
\hline
Random & 2306 \\
\hline
\end{tabular}
\end{center}
\end{table}
We again train a GNN on the navigation dataset but use the expected completion time of a given algorithm as the target instead of the one-hot label indicating the fastest algorithm. The network architectures used are identical to the previous experiments but with the softmax layer replaced with a single linear neuron. Both a GNN and a occupancy-grid-based fully-connected network were trained for each of the four algorithms. To test the predictions we sampled between $2$ and $10$ problems from the test set and used the neural predictors to predict the problem with fastest solution. We repeated this for $1000$ iterations and added up the actual expected completion time of each prediction to get a total expected completion time for all $1000$ predictions. We compared this to the total expected completion time using several other predictors. As baseline predictors we used a perfect predictor which always predicts the fastest problem, a random predictor which selects a random problem, a naive predictor which selects the problem with the fewest obstacles (if there are multiple problems with the minimum number of obstacles then a random one is chosen). 

The results for each algorithm are displayed in Table~\ref{table_random_problem_2D}
The results indicate that neural networks, and in particular, the GNN, are able to learn the relationship between problem structure and expected run time well enough to make useful predictions about the fastest problem among a set of test problems. We note that, while both neural networks are able to significantly reduce the total expected completion time over both the naive and random predictors, the GNN outperforms the fully-connected network. In spite of the opaque  nature of the dataset the neural models are able to predict the relative difficulty of problems. Next, we explore whether this learned problem difficulty can be leveraged to reduce the need for replanning in the task sequencing phase of TAMP.

\subsection{Navigation: TAMP}
   \begin{figure*}[t]
      \centering
      \begin{subfigure}{0.3\textwidth}
      \includegraphics[scale=0.4]{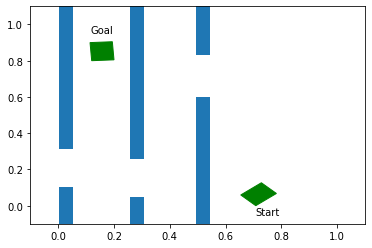}
      \caption{navigation element 1}
   \end{subfigure}
 \begin{subfigure}{0.3\textwidth}
\includegraphics[scale=0.84]{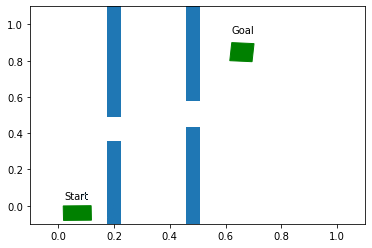}
\caption{navigation element 2}
\end{subfigure}
\begin{subfigure}{0.3\textwidth}
\includegraphics[scale=0.84]{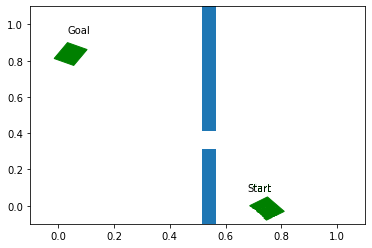}
\caption{navigation element 3}
\end{subfigure}
\caption{An example problem triple with valid transitions according to the TAMP constraints. Note that the x-interval and orientation quadrant must match between navigation elements but the y coordinate does not.}
\label{fig:2D_TMPExamp}
\end{figure*}

\begin{figure}[b]
      \centering
      
      \includegraphics[scale=0.5]{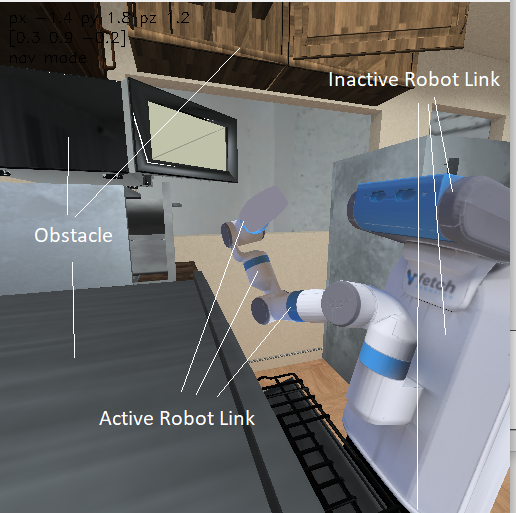}
      \caption{The iGibson fetch-griper robot in a kitchen environment with a selection of the objects labelled by type. Motion planning problems involve moving the gripper from positions like the one shown in this figure to other random positions}
      \label{fig:igibson}
\end{figure}
In this experiment we again test the ability of a GNN to predict motion planning problems that are expected to be solved quickly but rather than choosing the fastest amongst an arbitrary set of test problems we carry this experiment out in a more realistic TAMP setting.

A proof-of-concept TAMP validation in the navigation setting is constructed by asking for a three-element composite motion sequence of primitive elements with constraints restricting the sequencing of elements. The composite motion is specified with an initial condition defining a range of allowed starting x-positions and rotation angles and a goal state specifying a range of required x-positions and rotation angles. Additionally, sequencing constraints specify which navigation elements can follow one another. To model the kinds of constraints that would arise in real TAMP problems we partition the x-interval $[0,0.9]$ into $9$ equal segments, $[0,0.1),[0.1,0.2),...,[0.8,0.9]$ and the rotation angle space, $[-\pi,\pi]$ into quadrants, $[-\pi,-\pi/2),[-\pi/2,0),[0,\pi/2),[\pi/2,\pi)$. The initial conditions are then equivalent to a required starting (x-segment, quadrant) pair for the first navigation element and a required terminal (x-segment, quadrant) pair for the final (third) navigation element. The sequencing constraints require that the final (x-segment, quadrant) of one navigation element is the same as the starting (x-segment, quadrant) of the following navigation element. For example, an initial condition might specify: element 1 start $= ([0.7,0.8),[0,\pi/2))$, element 3 goal $=([0,0.1),[-\pi,-\pi/2))$ and a valid triple could satisfy, element 1, start $=([0.7,0.8),[0,\pi/2))$, goal $=([0.1,0.2),[-\pi,-\pi/2))$, element 2, start $=([0.1,0.2),[-\pi,-\pi/2))$, goal $=([0.6,0.7), [-\pi/2,0))$, and element 3, start $=([0.6,0.7),[-\pi/2,0))$, goal $=([0,0.1),[-\pi,-\pi/2))$. Fig.~\ref{fig:2D_TMPExamp} illustrates a feasible triple.
We tested our predictors on this problem by seeing if we can approximate the runtime of a navigation element triple satisfying the sequencing constraints. To do so we use our test set as the collection of potential navigation elements, this is the same environment used in the previous problem consisting of 1-3 narrow gaps. We randomly generate an initial condition and return all problem triples from the test set satisfying the initial and transition conditions. From this set the triple with the fastest expected completion time using RRTConnect is identified (using the information recorded in the test set). We then used various heuristics (learned GNN, learned fully-connected, and least obstacles) to predict the fastest elements within each triple to assemble the composite triple. This prediction of total composite runtime can then be compared with the minimum fastest possible triple identified. We repeated the process 100 times and took the sum to record the excess time taken by the heuristics. 

Table~\ref{table_TAMP_problem_2D} shows the results. We see that using neural networks to predict problem runtime usually identifies near-optimal triple sequences thereby greatly reducing the time taken to solve the TAMP problem.
\begin{figure}[b]
\centering
\includegraphics[scale=0.5]{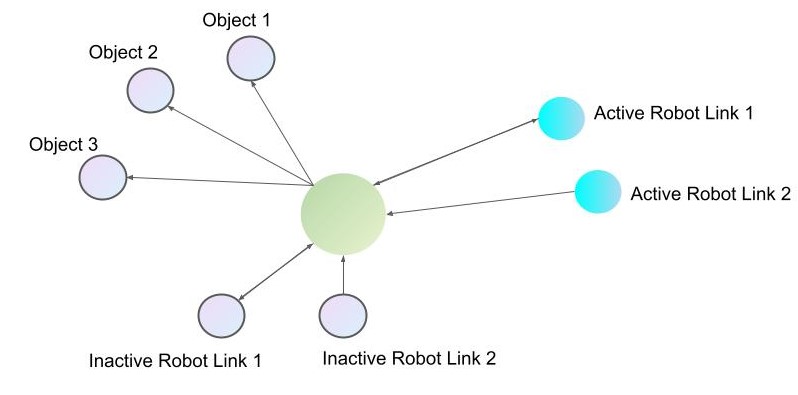}

\caption{Graph connections of a single node in an example problem of our 3D dataset. The 4 nodes with the closest positions to the example node are its neighbours. It is the neighbour of an additional 5 nodes whose neighbours are also determined using the k-nearest neighbours method }
\label{fig:node_con}
\end{figure}
\subsection{Manipulation: problem prediction}
\label{sec:manipulation}
For our final experiment we are again predicting the fastest planning problem amongst a test set but instead of a simple 2D navigation problem, we are testing the ability of a GNN to predict fast motion planning problems in a complex 3D environment.

Moving from a 2D navigation environment to a 7-DOF manipulation task in 3D environments would normally involve a large increase in the number of required parameters for a neural network to learn the problem. In particular, occupancy grid representations of the workspace result in an increase in the number of input neurons by an order of magnitude or more. In contrast, by using GNNs we are able to represent manipulation planning problems assembled from nodes with a comparable number of input features to the 2D navigation problems. In the 3D case, objects are represented by nodes with features including its 3D position (3 parameters), quaternion representation of orientation (4 parameters), the height, width, and length of its axis-aligned bounding box (3 parameters), and a 3 dimensional 1-hot label indicating object type. This gives nodes with 13 features versus the 8 features used for nodes in the 2D problem. We use this fact to test the ability of GNNs to predict the expected completion time of an algorithm on 3D motion planning problems. To accomplish this, we use a dataset generated using the iGibson environment \cite{shen2021igibson}.
We used the \textit{fetch-motion-planning.yaml} configuration of iGibson. In this  environment, a fetch-gripper robot with a 7 degree of freedom(DOF) arm is placed inside a house with several rooms including a kitchen and a living room. Various objects are contained within the rooms, such as cabinets, dishwashers, and ovens (see Fig \ref{fig:igibson}). We place the robot in a number of positions that are adjacent to a sufficient number of obstacles (usually between 6 and 11). From there, we randomly selected an initial and goal joint configuration for the robot and used OMPL to solve the problem with RRTConnect 40 times and recorded the average completion time. Our input graph consisted of nodes for each object in the vicinity of the robot including the links of the robot. Unlike the 2D problem, in the 3D problem we only include the initial position of the robot as nodes in the input graph. Because most robot links are inactive, representing both the final and initial position of each link as graph nodes results in a number of nodes with identical features.
\begin{figure}[b]
\centering
\includegraphics[scale=0.8]{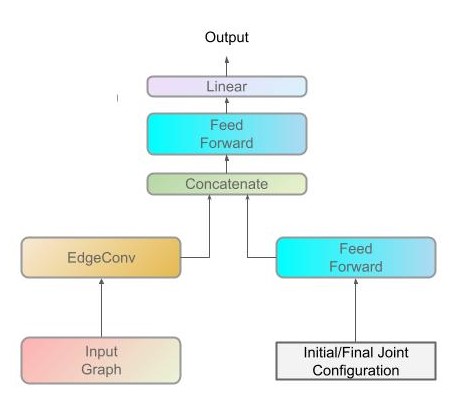}
\caption{Network architecture for the the 7 degree-of-freedom arm manipulation planning dataset. The EdgeConv node is used to represent the EdgeConv layers with architecture represented in Fig ~\ref{fig:EC_arch} }
\label{fig:nn_arch}
\end{figure}
We therefore only include the initial position of the robot links in the workspace as nodes in the graph. In order to capture the full problem description in the input, the initial and final joint positions of the robot were included in addition to the graph as inputs. The GNN component is identical to the previous problem with each node in the graph connected to it's 4 nearest neighbours (see Fig. ~\ref{fig:node_con} for an example of the connections for a node in a given problem). The full network thus involves a GNN to process the input graph, a fully-connected network to process the initial and final joint configurations, and a final fully-connected network which takes the output of the previous two networks and produces the final output (see Fig.~\ref{fig:nn_arch}). The features of the graph nodes are as listed above with the one-hot label indicating whether the object is an obstacle, an active robot link (one that is involved in the motion planning problem), or an inactive robot link. The fully connected network consisted of 2 layers of $512$ and $256$ hidden units, and the final fully connected layer was $1000$ hidden units. The training dataset consisted of 7000 examples. 
The network was tested as before by using it to predict the fastest problem among a selection of problems from the test set. This was repeated 1000 times and the total expected time for the predictions was recorded and compared to predictions made by a perfect predictor, a least-obstacles predictors, and a random predictor. The fully-connected occupancy-grid predictor is not shown as it does not scale to this problem. 

The results are found in table \ref{table_random_problem_3D}. The motion planning problems generated in iGibson were more difficult than the navigation problems and a timeout of 10 seconds was used. Again, the GNN prediction accurately assesses problem difficulty. This demonstrates the potential of runtime predictors for use in TAMP frameworks such as PDDLstreams \cite{Garrett2020PDDLStreamIS} in realistic planning domains. 

\section{CONCLUSIONS AND FUTURE WORK}

We have demonstrated the ability of neural networks to capture (though offline training) the structure of motion planning problems sufficient to predict the likely runtime of given SBMP algorithms. GNNs are particularly well-suited and show better predictive accuracy and scaling with problem size than voxelized representations of the environment. Accurate runtime prediction is then leveraged to demonstrate the promise of planner portfolios which dynamically select the planner which is a best match to a particular planning problem. Planning portfolios may be very useful in the compute-bound settings common to many robotic platforms. Moreover, through a simplified proof-of-concept we have shown that runtime prediction might also accelerate integrated task and motion planning of complex motions by identifying more easily solved motion planning subproblems.

Our results and experiments  clearly demonstrate the potential of harnessing the learning power of GNNs for extracting the relationship between a motion planning problem and the SBMP algorithm used to solve the problem. This is an important first step towards future work that must assess the computational benefits in practical settings. The application to task and motion planning is particularly interesting given the difficulty of TAMP. Many interesting questions remain regarding this use-case; most prominently, our simulated TAMP problem did not explore the possible role that expected runtimes might serve in heuristically guiding search over task plans.

Quantification of the benefits of the relational structure GNNs should be explored more fully. How well do the predictions learned from one  workspace transfer to markedly different workspace geometries? The compositional nature of GNNs means that they should be better suited to this transfer learning than fully-connected networks but this requires further validation. The direct representation of the workspace as a graph provides direct insight into the interpretability of GNN runtime predictors. As an example, removing a node corresponding to an obstacle and rerunning the network allows for quantification of the effect of the obstacle on runtime. Recent work on explainability methods for GNNs (see \cite{graphexplain} for a survey) may offer further clues to
the operation of GNNs on workspace graphs and facilitate the design of architectures having improved generalizability.

Lastly, we note that all our experiments focused on the training of networks to learn the runtime of a given \textit{single} robot. This is because the input to the neural network only contains information about the environment and the motion planning problem but not the structure of the robot itself. Future work might include robot-specific information such as joint type into robot nodes thereby allowing the network to learn the relationship between the structure of the robot \textit{and} the speed of an SBMP for a given motion planning problem.

\addtolength{\textheight}{-12cm}   





\section*{ACKNOWLEDGMENT}

This research has been funded by a series of MITACS grants: FR59629, FR59630, FR70699, FR70999, and FR71000.

\bibliographystyle{IEEEtran}
\bibliography{IEEEabrv,mybibfile}

\begin{thebibliography}{10}
\providecommand{\url}[1]{#1}
\csname url@rmstyle\endcsname
\providecommand{\newblock}{\relax}
\providecommand{\bibinfo}[2]{#2}
\providecommand\BIBentrySTDinterwordspacing{\spaceskip=0pt\relax}
\providecommand\BIBentryALTinterwordstretchfactor{4}
\providecommand\BIBentryALTinterwordspacing{\spaceskip=\fontdimen2\font plus
\BIBentryALTinterwordstretchfactor\fontdimen3\font minus
  \fontdimen4\font\relax}
\providecommand\BIBforeignlanguage[2]{{%
\expandafter\ifx\csname l@#1\endcsname\relax
\typeout{** WARNING: IEEEtran.bst: No hyphenation pattern has been}%
\typeout{** loaded for the language `#1'. Using the pattern for}%
\typeout{** the default language instead.}%
\else
\language=\csname l@#1\endcsname
\fi
#2}}

\bibitem{essentials}
S.~LaValle, ``Motion planning: The essentials,'' \emph{{IEEE} Robot. Automat.
  Mag.}, vol.~18, pp. 79--89, Mar. 2011.

\bibitem{RRT}
S.~M. LaValle, ``Rapidly-exploring random trees: A new tool for path
  planning,'' Iowa State Univ., Tech. Rep., Oct. 1998.

\bibitem{RRTConnect}
J.~Kuffner and S.~LaValle, ``{RRT}-connect: An efficient approach to
  single-query path planning,'' in \emph{Proc. Millennium Conf. {IEEE} Int.
  Conf. Robot. Automat.}, vol.~2, April 2000, pp. 995--1001.

\bibitem{journals/trob/KavrakiSLO96}
L.~E. Kavraki, P.~Svestka, J.-C. Latombe, and M.~H. Overmars, ``Probabilistic
  roadmaps for path planning in high-dimensional configuration spaces,''
  \emph{IEEE Trans. Robotics Autom.}, vol.~12, no.~4, pp. 566--580, 1996.

\bibitem{shen2021igibson}
B.~Shen, F.~Xia, C.~Li, R.~Mart\'in-Mart\'in, L.~Fan, G.~Wang,
  C.~Pérez-D'Arpino, S.~Buch, S.~Srivastava, L.~P. Tchapmi, M.~E. Tchapmi,
  K.~Vainio, J.~Wong, L.~Fei-Fei, and S.~Savarese, ``i{G}ibson 1.0: A
  simulation environment for interactive tasks in large realistic scenes,'' in
  \emph{Proc. {IEEE}/{RSJ} Int. Conf. Intell. Robots and Syst.}, Dec. 2021, pp.
  7520--7527.

\bibitem{DBLP:journals/corr/abs-2010-01083}
\BIBentryALTinterwordspacing
C.~R. Garrett, R.~Chitnis, R.~Holladay, B.~Kim, T.~Silver, L.~P. Kaelbling, and
  T.~Lozano-Pérez, ``Integrated task and motion planning,'' 2020. [Online].
  Available: \url{https://arxiv.org/abs/2010.01083}
\BIBentrySTDinterwordspacing

\bibitem{CVAE1}
B.~Ichter, J.~Harrison, and M.~Pavone, ``Learning sampling distributions for
  robot motion planning,'' in \emph{Proc. {IEEE} Int. Conf. Robot. Automat.},
  May 2018, pp. 7087--7094.

\bibitem{CVAE2}
S.~M. . R.~N. Liu, K., ``Learned sampling distributions for efficient planning
  in hybrid geometric and object-level representations,'' in \emph{Proc. IEEE
  Int. Conf. Robot. Automat.}, May 2020, pp. 9555--9562.

\bibitem{NGMP2}
A.~H. Qureshi and M.~C. Yip, ``Deeply informed neural sampling for robot motion
  planning,'' in \emph{Proc. {IEEE/RSJ} Int. Conf. Int. Robots Syst.}, Oct.
  2018, pp. 6582--6588.

\bibitem{DBLP:journals/trob/QureshiMSY21}
A.~H. Qureshi, Y.~Miao, A.~Simeonov, and M.~C. Yip, ``Motion planning networks:
  Bridging the gap between learning-based and classical motion planners,''
  \emph{{IEEE} Trans. Robot.}, vol.~37, no.~1, pp. 48--66, Feb. 2021.

\bibitem{DBLP:conf/iros/SungKL21}
Y.~Sung, L.~P. Kaelbling, and T.~Lozano{-}P{\'{e}}rez, ``Learning when to quit:
  Meta-reasoning for motion planning,'' in \emph{Proc. {IEEE/RSJ} Int. Conf.
  Int. Robots Syst.}, Sept. 2021, pp. 4692--4699.

\bibitem{RRT*}
S.~Karaman and E.~Frazzoli, ``Sampling-based algorithms for optimal motion
  planning,'' \emph{The Int. J. Robot. Research}, vol.~30, no.~7, pp. 846--894,
  June 2011.

\bibitem{10.5555/2832747.2832840}
F.~Hutter, L.~Xu, H.~H. Hoos, and K.~Leyton-Brown, ``Algorithm runtime
  prediction: Methods \& evaluation,'' \emph{Artif. Intell.}, vol. 206, pp.
  79--111, Jan. 2014.

\bibitem{GNN1}
M.~Gori, G.~Monfardini, and F.~Scarselli, ``A new model for learning in graph
  domains,'' in \emph{Proc. {IEEE} Int. Joint Conf. Neural Networks}, vol.~2,
  Jul. 2005, pp. 729--734.

\bibitem{GNN2}
Y.~Li, R.~Zemel, M.~Brockschmidt, and D.~Tarlow, ``Gated graph sequence neural
  networks,'' in \emph{Proc. Int. Conf. Learn. Representations}, April 2016.

\bibitem{DBLP:conf/iclr/XuHLJ19}
\BIBentryALTinterwordspacing
K.~Xu, W.~Hu, J.~Leskovec, and S.~Jegelka, ``How powerful are graph neural
  networks?'' in \emph{Proc. Int. Conf. Learn. Representations}, May 2019.
  [Online]. Available: \url{https://openreview.net/forum?id=ryGs6iA5Km}
\BIBentrySTDinterwordspacing

\bibitem{zhou2018graph}
\BIBentryALTinterwordspacing
J.~Zhou, G.~Cui, Z.~Zhang, C.~Yang, Z.~Liu, and M.~Sun, ``Graph neural
  networks: A review of methods and applications,'' 2018. [Online]. Available:
  \url{http://arxiv.org/abs/1812.08434}
\BIBentrySTDinterwordspacing

\bibitem{DBLP:journals/tog/WangSLSBS19}
Y.~Wang, Y.~Sun, Z.~Liu, S.~E. Sarma, M.~M. Bronstein, and J.~M. Solomon,
  ``Dynamic graph {CNN} for learning on point clouds,'' \emph{{ACM} Trans.
  Graph.}, vol.~38, no.~5, pp. 1--12, Oct. 2019.

\bibitem{GNNmp}
\BIBentryALTinterwordspacing
A.~Khan, A.~Ribeiro, V.~Kumar, and A.~G. Francis, ``Graph neural networks for
  motion planning,'' 2020. [Online]. Available:
  \url{https://arxiv.org/abs/2006.06248}
\BIBentrySTDinterwordspacing

\bibitem{PyTorch}
A.~Paszke, S.~Gross, F.~Massa, A.~Lerer, J.~Bradbury, G.~Chanan, T.~Killeen,
  Z.~Lin, N.~Gimelshein, L.~Antiga, A.~Desmaison, A.~Kopf, E.~Yang, Z.~DeVito,
  M.~Raison, A.~Tejani, S.~Chilamkurthy, B.~Steiner, L.~Fang, J.~Bai, and
  S.~Chintala, ``Pytorch: An imperative style, high-performance deep learning
  library,'' in \emph{Advances Neural Inf. Process. Syst.}, Dec. 2019, pp.
  8024--8035.

\bibitem{PyTorchgeo}
\BIBentryALTinterwordspacing
M.~Fey and J.~E. Lenssen, ``Fast graph representation learning with pytorch
  geometric,'' 2019. [Online]. Available: \url{http://arxiv.org/abs/1903.02428}
\BIBentrySTDinterwordspacing

\bibitem{trrt}
L.~Jaillet, J.~Cortés, and T.~Siméon, ``Sampling-based path planning on
  configuration-space costmaps,'' \emph{{IEEE} Trans. Robotic.}, vol.~26, pp.
  635 -- 646, Sept. 2010.

\bibitem{Bohlin01arandomized}
R.~Bohlin and L.~E. Kavraki, ``A randomized approach to robot path planning
  based on lazy evaluation,'' in \emph{Handbook on Randomized Computing}, 2001,
  pp. 221--249.

\bibitem{OMPL}
I.~A. {\c{S}}ucan, M.~Moll, and L.~E. Kavraki, ``The open motion planning
  library,'' \emph{{IEEE} Robotic. Automat. Mag.}, vol.~19, no.~4, pp. 72--82,
  December 2012.

\bibitem{Garrett2020PDDLStreamIS}
C.~R. Garrett, T.~Lozano-Perez, and L.~P. Kaelbling, ``{PDDL}stream:
  Integrating symbolic planners and blackbox samplers via optimistic adaptive
  planning,'' in \emph{Proc. Int. Conf. Automat. Planning Scheduling}, May
  2020, pp. 440--448.

\bibitem{graphexplain}
\BIBentryALTinterwordspacing
H.~Yuan, H.~Yu, S.~Gui, and S.~Ji, ``Explainability in graph neural networks: A
  taxonomic survey,'' 2020. [Online]. Available:
  \url{https://arxiv.org/abs/2012.15445}
\BIBentrySTDinterwordspacing

\end{thebibliography}

\end{document}